\documentclass{article}

\usepackage{arxiv}

\usepackage[utf8]{inputenc} 
\usepackage[T1]{fontenc}    
\usepackage{hyperref}       
\usepackage{url}            
\usepackage{booktabs}       
\usepackage{amsfonts}       
\usepackage{amsmath}       
\usepackage{nicefrac}       
\usepackage{microtype}      
\usepackage{lipsum}		
\usepackage{graphicx}
\usepackage[square,sort,comma,numbers]{natbib}
\usepackage{doi}

\title{3D scene reconstruction from monocular spherical video with motion parallax}

\author{ \href{https://orcid.org/0000-0001-5366-1505}{\includegraphics[scale=0.06]{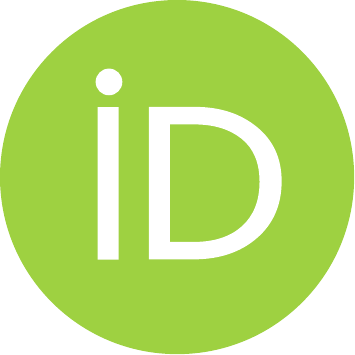}\hspace{1mm}Kenji Tanaka}\\
	Fprime Project\\
	\texttt{tanaken@computer.org} \\
}

\renewcommand{\shorttitle}{3D scene reconstruction from monocular spherical video with motion parallax}

\hypersetup{
pdftitle={3D scene reconstruction from monocular spherical video with motion parallax},
pdfsubject={cs.CV},
pdfauthor={Kenji Tanaka},
pdfkeywords={computer vision, 3D reconstruction, spherical depth information},
}

\begin{document}
\maketitle

\begin{abstract}
In this paper, we describe a method to capture nearly entirely spherical (360 degree) depth information using two adjacent frames from a single spherical video with motion parallax.
After illustrating a spherical depth information retrieval using two spherical cameras, we demonstrate monocular spherical stereo by using stabilized first-person video footage. 
Experiments demonstrated that the depth information was retrieved on up to 97\% of the entire sphere in solid angle. At a speed of 30 km/h, we were able to estimate the depth of an object located over 30 m from the camera.
We also reconstructed the 3D structures (point cloud) using the obtained depth data and confirmed the structures can be clearly observed.
We can apply this method to 3D structure retrieval of surrounding environments such as 1) previsualization, location hunting/planning of a film, 2) real scene/computer graphics synthesis and 3) motion capture.
Thanks to its simplicity, this method can be applied to various videos.
As there is no pre-condition other than to be a 360 video with motion parallax, we can use any 360 videos including those on the Internet to reconstruct the surrounding environments. The cameras can be lightweight enough to be mounted on a drone. We also demonstrated such applications.
\end{abstract}


\section{Introduction}
Capturing 360-degree information of the surrounding environments and objects plays an important role in many applications, including vision-based communication, automatic robot or vehicle navigation, geographical information services and motion capture for filmmakers.
While an immersive video is still two-dimensional (2D), three-dimensional (3D) environment capture usually needs special sensing devices. This paper describes the retrieval of 3D information from a monocular spherical video with motion parallax.

\subsection{Related works}

One approach to capture the 3D structure of the surrounding environments is to use other sensing systems than ordinary cameras, such as a laser range finder or an infrared camera and a retroreflective ball, and then to fuse the information into the vision data.
Google's Street View vehicle has a laser range finder to capture the entire 3D world \cite{anguelov2010google}.
The special hardware can be very useful if it is affordable and performs adequately.

Another approach is vision-based capture.
Many efforts have been made to collect more ray information than is obtainable from a normal perspective camera.
The use of multiple cameras is one option to capture full-view spherical stereoscopic images \cite{tanhashi2001acquisition}.
360Heros and Google Jump capture 360 video using multiple cameras. To retrieve depth information from these multiple cameras, sufficient bandwidth and computing power to calculate stereo pairs are required.
Although they are not full-view spherical capture, curved mirrors and reflective optics have also been used to capture wide-angle or 360-degree depth and environment \cite{shimamura2000construction} \cite{lin2003high} \cite{tanaka2005tornado}.
Broxton et al. \cite{broxton2020immersive} used 46 cameras on the surface of a hemispherical dome to reconstruct immersive light field video. In order to reduce the data size, they are using a fixed number of RGBA+depth layers.

Another approach to retrieve 3D information from a 2D video is the use of multiple frames of a single video sequence.
Bartczak et al. \cite{bartczak2007extraction} used an omnidirectional camera to reconstruct 3D surface models from a video sequence. They used the concept of Structure from Motion and required multiple frames to reconstruct the 3D environment.
Odometry studies \cite{scaramuzza2008appearance} \cite{munteanu2014visual} have also been conducted.
These studies are mainly focused on two-dimensional motions of the viewpoint on the ground and also require multiple frames.

The development of a high-resolution entire 360 camera such as Ricoh THETA and Insta 360 widened the possibility of capturing the environment in an additional way.
There have been several studies to reconstruct 3D data using multiple spherical camera or spherical stereo pairs.
Li et al. \cite{li2006real} \cite{li2008binocular} proposed binocular spherical stereo as an epipolar geometry for spherical stereo pairs and found the effectiveness of using fish-eye cameras, while
Ma et al. \cite{ma20153d} applied this concept to an actual spherical camera pair. 
Kim et al. \cite{kim20133d} proposed a 3D environment modeling method using multiple pairs of high-resolution spherical images.
These studies used multiple or at least two spherical cameras.
Bodington et al. \cite{bodington2015rendering} proposed a technique to reconstruct an omnistereo image pair from two spherical images captured by a known vertical displacement. 
Their efforts are mainly focused on the reconstruction of omnistereo imagery.
Recent studies include creating 6 degree-of-freedom (DoF) views from a single 360 camera \cite{huang20176} and machine learning-based depth estimation from a single spherical image \cite{zioulis2018omnidepth}. While the former focuses on a 6-DoF view synthesis for VR headsets, the latter is mainly for indoor images.

First, we extend the concept of binocular spherical stereo to capture the 3D geometry and reconstruct the point cloud by using spherical images from only two positions separated vertically.
Then, we propose a method to estimate depth information using two frames from a spherical video sequence with motion parallax.

\section{Binocular spherical stereo}
\label{sec:binocular_spherical_stereo}

In this section, we describe the method and results of the nearly entirely spherical depth estimation from binocular sphere images.

\subsection{Method}
\label{subsec:binocular_method}
An ideal entirely spherical camera maps the incoming light from every direction to its central point $C$ (we call this a viewpoint) on a sphere centered at $C$.
An equirectangular image is obtained directly from this spherical information and now is the defacto format for spherical images.

\begin{figure}[ht]
  \centering
  $\vcenter{\hbox{\includegraphics[height=1.8in]{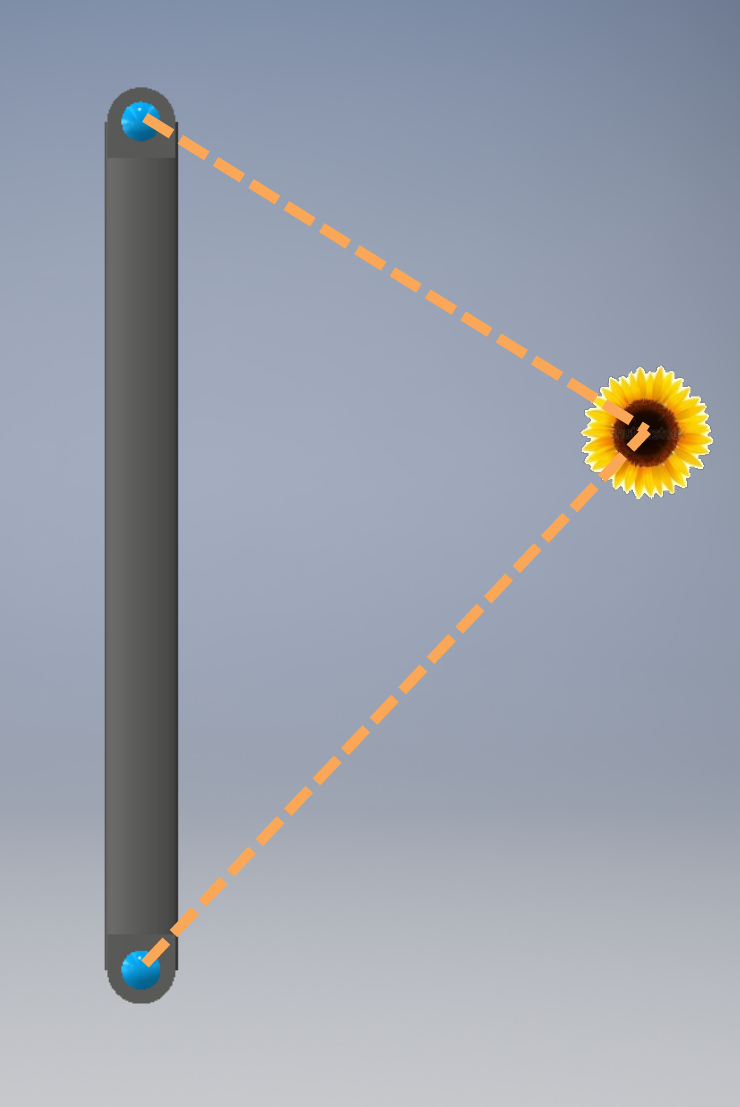}}}$
  \hspace*{1.5em}
  $\vcenter{\hbox{\includegraphics[height=1.3in]{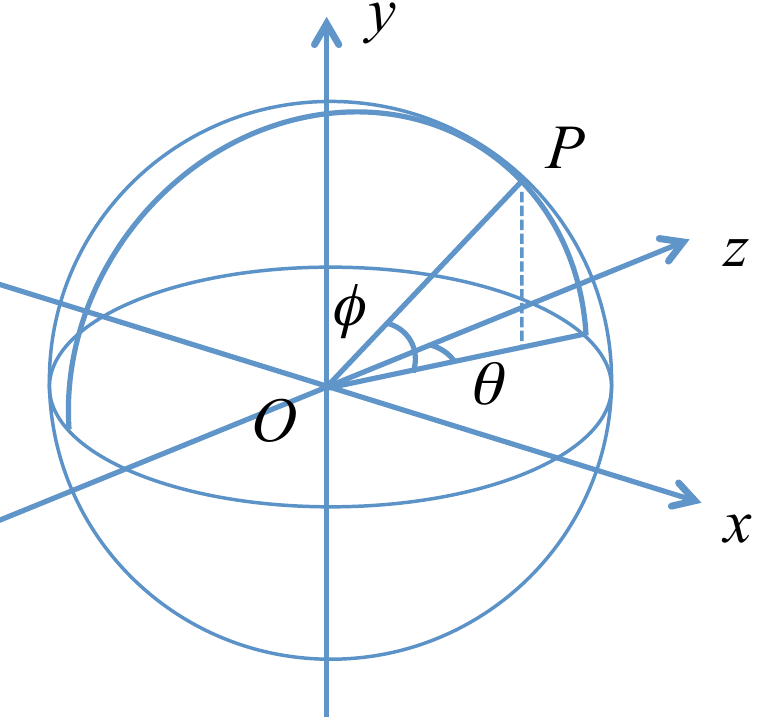}}}$
  \caption{Binocular spherical stereo camera configuration (left) and coordinate system for each optics (right).}
  \label{fig:binocular_cameras}
\end{figure}

\begin{figure}[ht]
  \centering
  \includegraphics[width=3.0in]{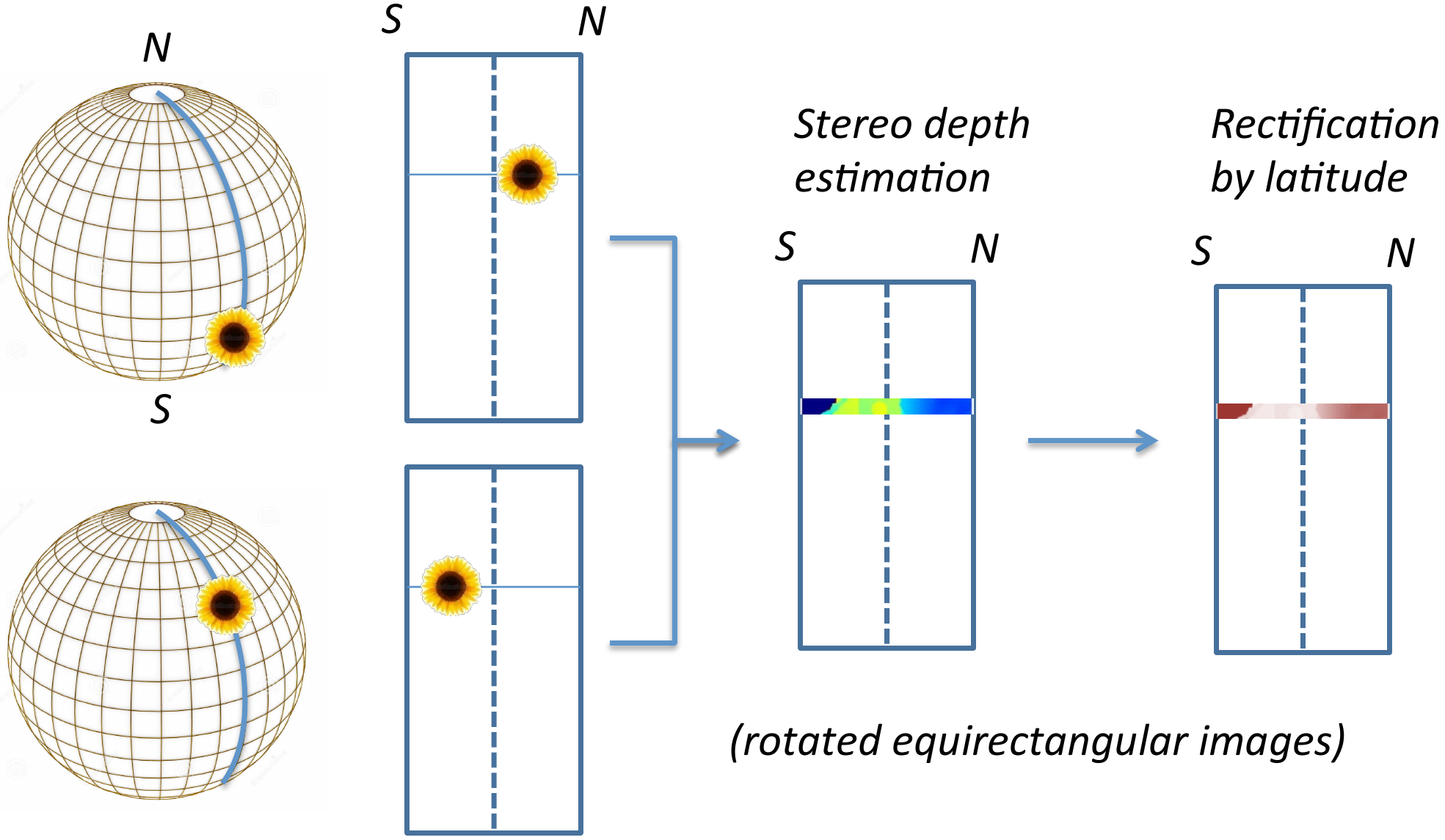}
  \caption{Principle of binocular spherical stereo. When seen from two positions separated vertically, an object is projected onto the same meridian. When we rotate these images by 90$^{\circ}$ clockwise, we can obtain two images with horizontal parallax.
  }
  \label{fig:twoviewpoints}
\end{figure}

Figure \ref{fig:binocular_cameras} shows two such cameras placed vertically along with the coordinate system.
Here, binocular spherical stereo is formulated as the following.
When seen from two positions separated vertically, an object is projected onto the same meridian in the upper and lower images (Figure \ref{fig:twoviewpoints}).
When we rotate these images by 90$^{\circ}$ clockwise, we can obtain two images with horizontal parallax. Once we obtain these images, we can apply any stereo algorithm.
In this configuration, the epipole for the upper viewpoint is the south pole and the epipole for the lower viewpoint is the north pole\footnote{Here we use the analogy of earth to address geometry on a sphere.}. This means that all of the meridians are the epipolar lines.

The distance of an object that has $n$-pixel parallax is calculated as follows:
\begin{equation} \label{eq:D_npixels}
D_{n} = \frac{d}{\tan \left(\frac{n}{R_\text{vertical}} \pi \right)} \simeq \frac{d \cdot R_\text{vertical}}{\pi n}
\end{equation}
where $n$ is the distance between the point in the upper image and the corresponding point in the lower image in the number of pixels, $d$ is the distance of two cameras and $R_\text{vertical}$ is the vertical resolution of the equirectangular expression.
The last step is the latitude-based rectification.
As is described in Appendix \ref{sec:apparent_reduction}, we have to rectify the distance of an object based on the latitude $\phi$ of the incoming light.
\begin{equation} \label{eq:Rectification}
  D_{REAL} \simeq \frac{D_{APPARENT}}{\cos \phi}
\end{equation}

\begin{figure}[ht]
\centering
  \includegraphics[width=3.0in]{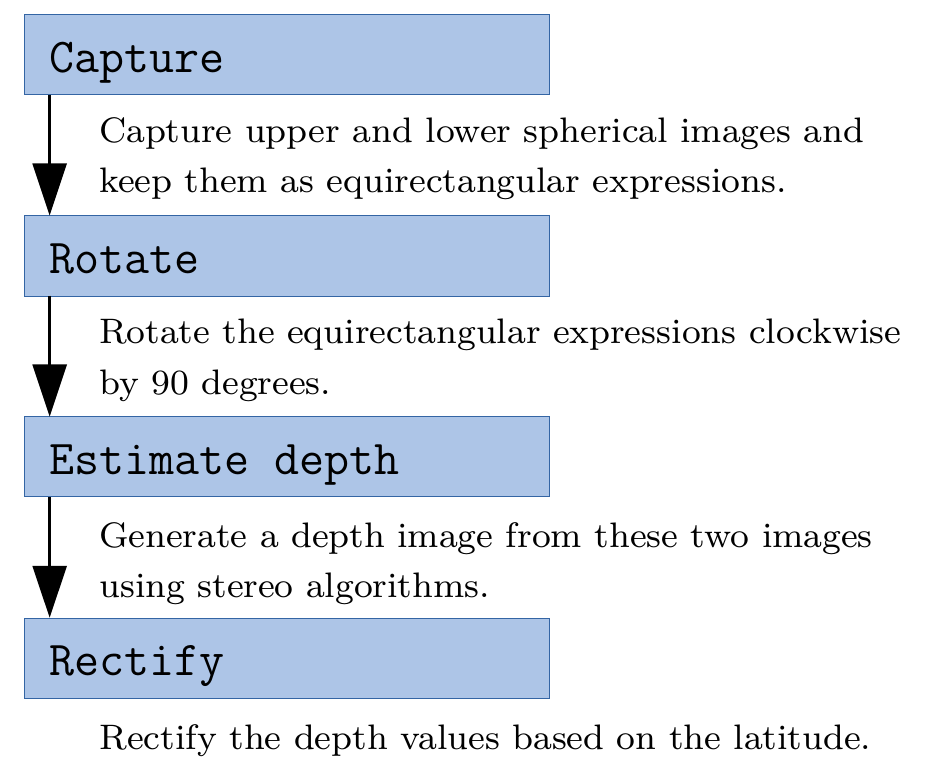}
  \caption{The processing pipeline of binocular spherical stereo.}
  \label{fig:binocular_spherical_stereo_pipeline}
\end{figure}

Figure \ref{fig:binocular_spherical_stereo_pipeline} shows the processing pipeline.

\subsection{Results}
\label{subsec:binocular_results}

\begin{figure}[ht]
  \centering
  \includegraphics[width=3.0in]{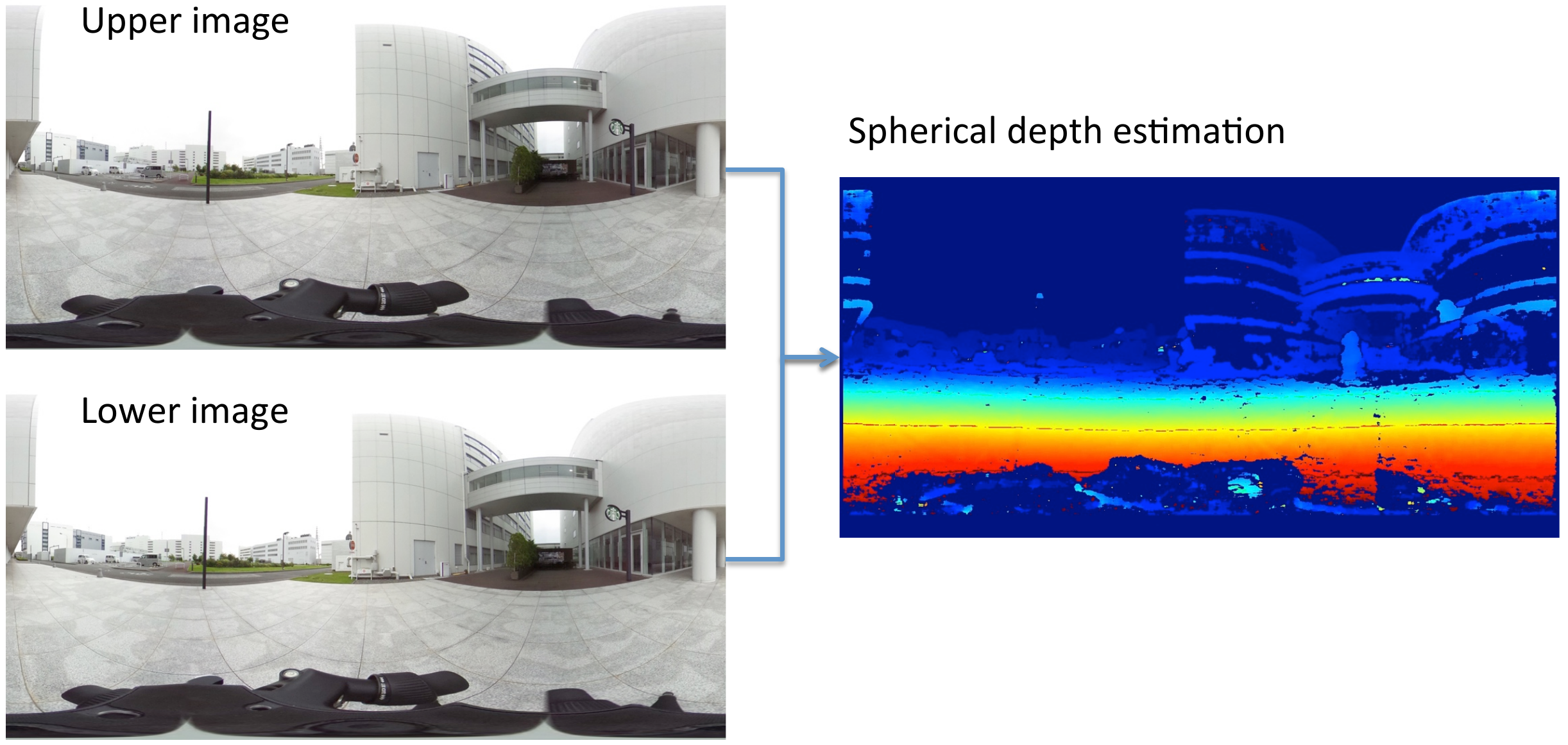}
  \caption{Binocular spherical stereo results. The images were shot from two positions vertically separated by 20 cm.
  }
  \label{fig:binocular}
\end{figure}

Figure \ref{fig:binocular} shows a color-mapped result of reconstructing depth information.
The upper and lower images were shot with a Ricoh THETA S from two positions vertically separated by 20 cm.
At this time, we captured the images by two separate exposures using a single camera.
The equirectangular image has the dimensions of 3,584$\times$1,792.
We used a block matching based stereo matcher ({\tt StereoBM} in OpenCV4) to estimate the depth. The depth is calculated as the reciprocal of the parallax (of a certain point), while we used the value of the parallax itself in the depth color map. We set the number of disparities at 96.
With this resolution, we can clearly see the edges of the building structure, the ground and the far buildings.
By using Equation \ref{eq:D_npixels}, we can calculate $D_{n \rightarrow 1} = 61 \text{[m]}$
near the horizon as $d = 0.2\text{[m]}$ and $R_\text{vertical} = 960$. We found that the farthest building is located over 120 m from the cameras. As $D_{n \rightarrow 0.5} = 122.2$, we can confirm that the stereo matching algorithm estimates the depth from half-pixel parallax.
We also found that we can capture depth information from the south latitude 80$^{\circ}$ (the pattern on the ground) to the north latitude 80$^{\circ}$ (the building structure).
This covers over 98.5\% of the entire solid angle of the sphere.
We found that the edge of the building is located 20 m from the cameras.

\begin{figure}[htbp]
  \centering
  \includegraphics[width=3.0in]{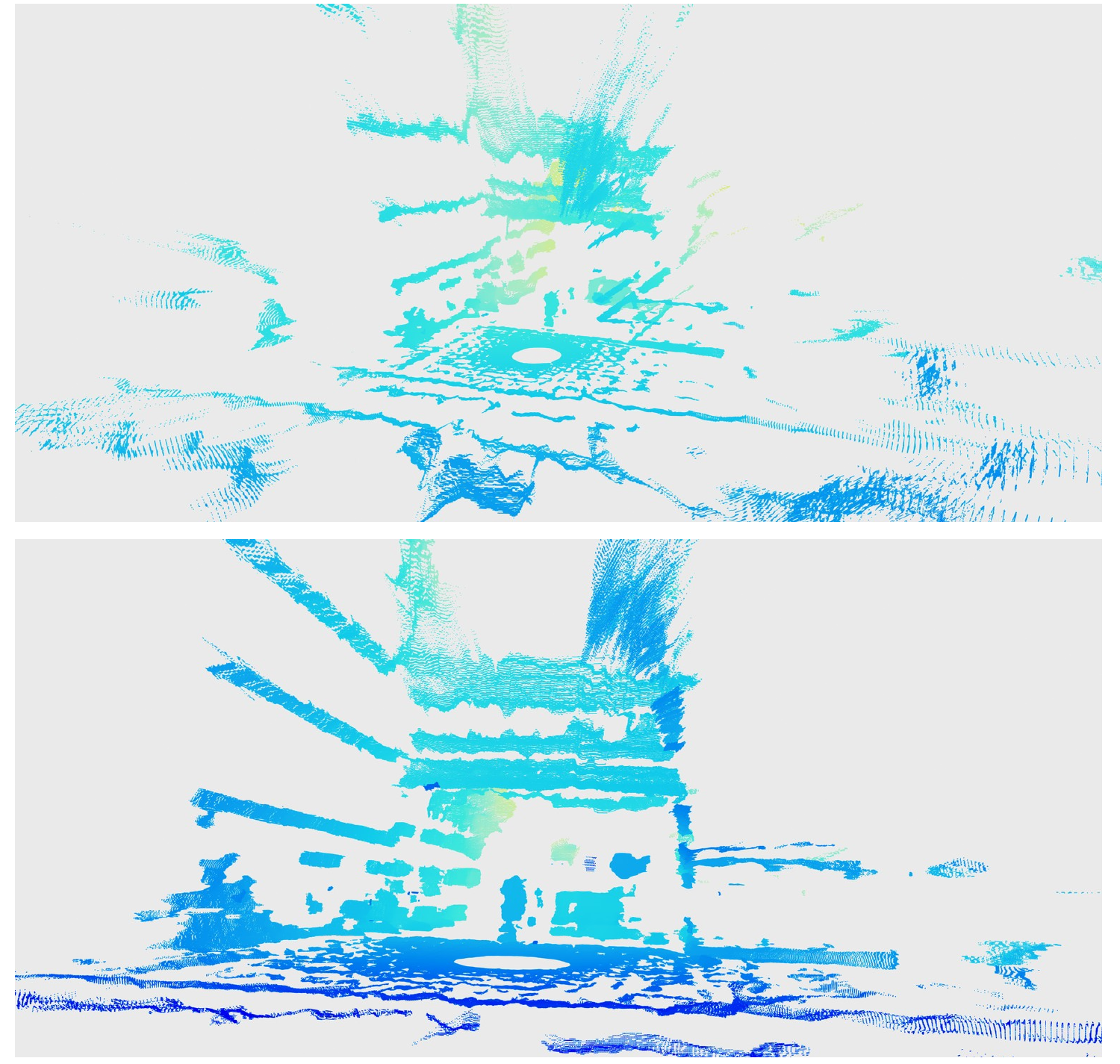}
  \caption{Point cloud reconstructed from the above data. Bird view showing the square, the road and the building (top) and
    View from the ground (bottom). The square ground and the corridor between two buildings are clearly observed.}
  \label{fig:binocular_pc}
\end{figure}

Figure \ref{fig:binocular_pc} shows the reconstructed point cloud.
For each pixel $(\theta, \phi)$ in the equirectangular depth map, we calculated a coordinate of a point $(r \sin(\theta) \cos(\phi), r \sin(\phi), - r \cos(\theta) \cos(\phi))$, where $r$ is the depth value.
We used Open3D\cite{zhou2018open3d} to visualize the data. We confirmed that the structures such as roads and buildings can be clearly observed.
We compared the top view with the actual floor plan and confirmed that the maximum error in the dimensions obtained from the point cloud is less than 5\% for the structure on the ground level whose distance from the camera is within 10 m.

\begin{figure}[ht]
  \centering
  \includegraphics[width=3.0in]{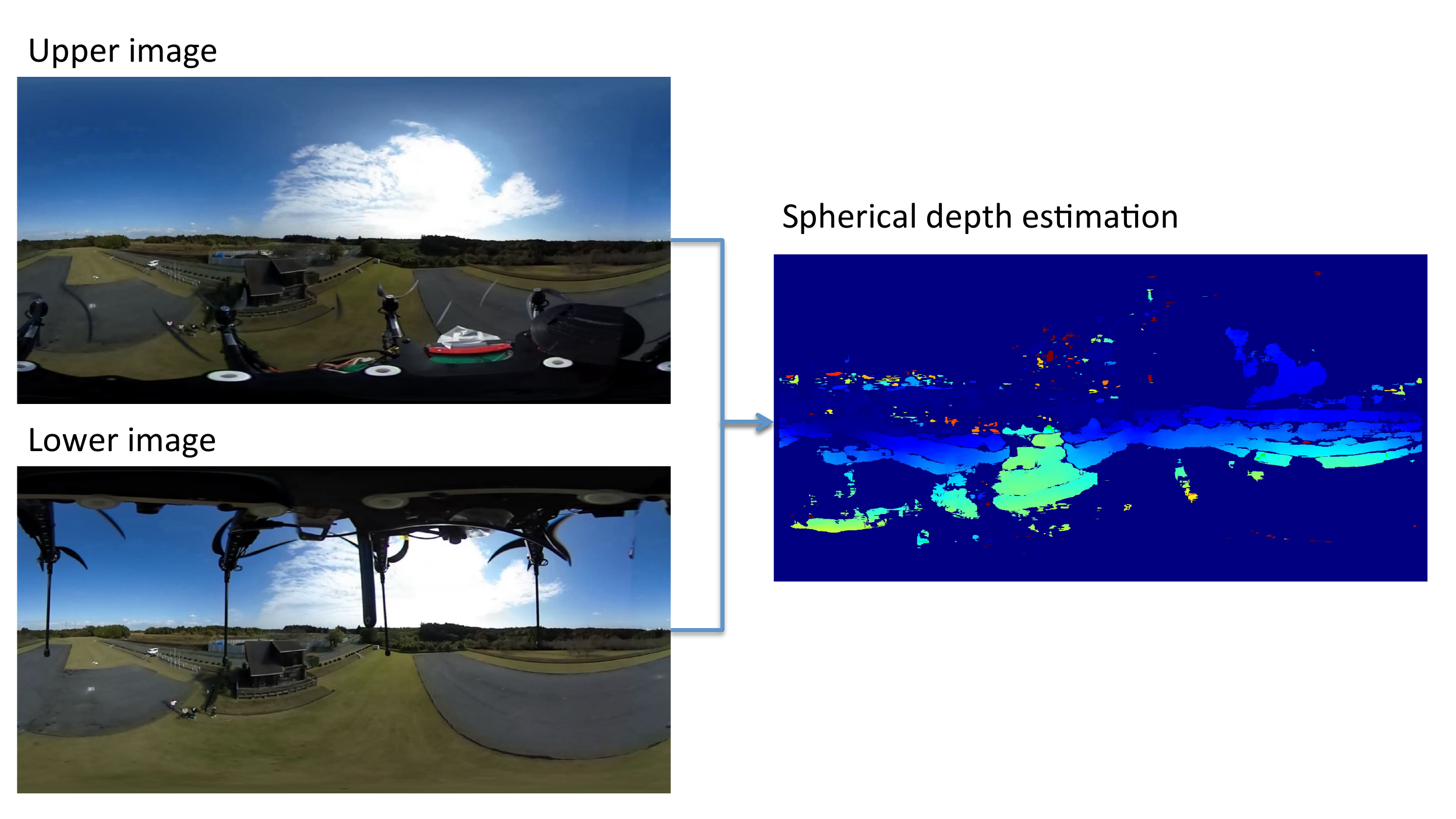}
  \caption{Example of spherical depth reconstruction from binocular spherical videos shot with two Ricoh THETA S cameras placed above and below a drone.}
  \label{fig:drone_shot}
\end{figure}

Figure \ref{fig:drone_shot} is an example of applying this method to aerial videos.
This binocular spherical stereo pair was shot with two Ricoh THETA S cameras attached above and below a drone. The resolution of the equirectangular video was 1,960$\times$960, and the video was recorded at 30 frames per second (FPS).
Two cameras are separated by 35 cm vertically. We can clearly see the ground and the structure. The small house is located over 100 m from the drone.

\paragraph{Alignment of two cameras}

\begin{figure}[ht]
  \centering
  \includegraphics[width=3.0in]{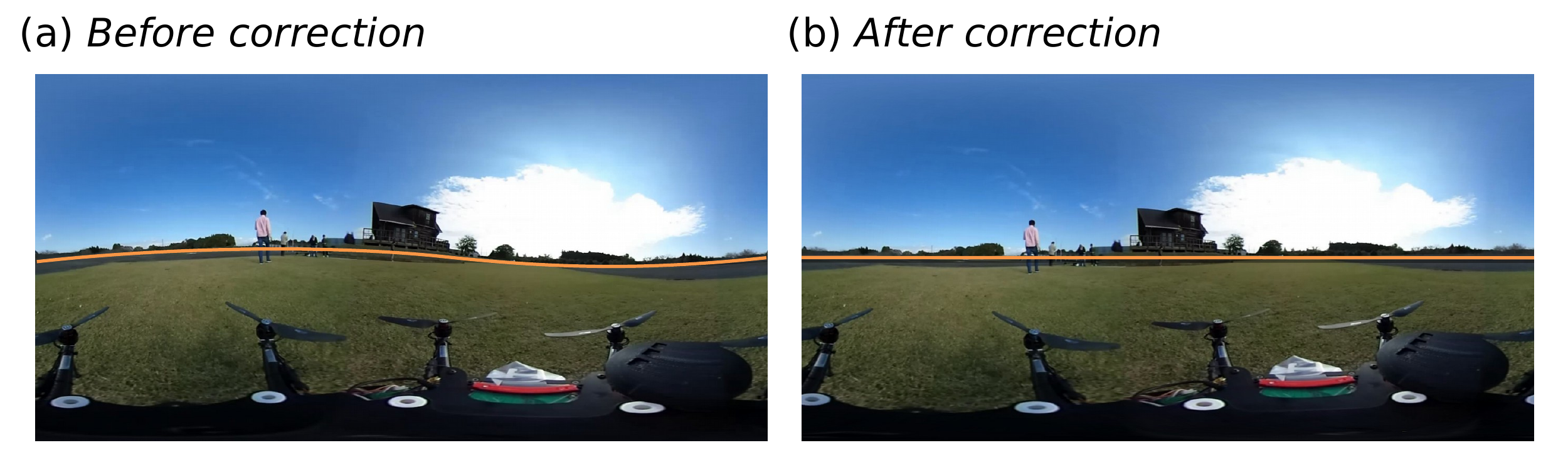}
  \caption{The deformation of a horizontal line and its correction.
  }
  \label{fig:rpycorrect_drawings}
\end{figure}

In a binocular spherical stereo, it is essential to align the coordinate system (in Figure \ref{fig:binocular_cameras}) so that the $x$, $y$ and $z$ axes of two cameras are in parallel.
Otherwise, the performance of the depth estimation will significantly deteriorate.
Figure \ref{fig:rpycorrect_drawings} shows how the misalignment of the upper camera causes the deformation of the horizontal line, when the bottom camera is aligned so that its $y$ axis becomes perpendicular to the ground.
Let $\alpha$, $\beta$ and $\gamma$ be the pitch, yaw and roll angles relative to those of the bottom camera respectively. Then the rotation around the $y$ axis ($\beta$) causes a horizontal shift in the top camera equirectangular image, while the rotation around the $x$ and $z$ axes ($\alpha$ and $\gamma$ respectively) cause a sinusoidal deformation of the horizontal curve.
We can cancel this deformation by identifying $\alpha$, $\beta$, $\gamma$ mainly by observing the pitch, yaw and roll fluctuations in the equirectangular images in the process of camera calibration.
The details of the correction can be found in Appendix \ref{sec:rpy_correct}.

\begin{figure}[ht]
  \centering
  \includegraphics[width=3.0in]{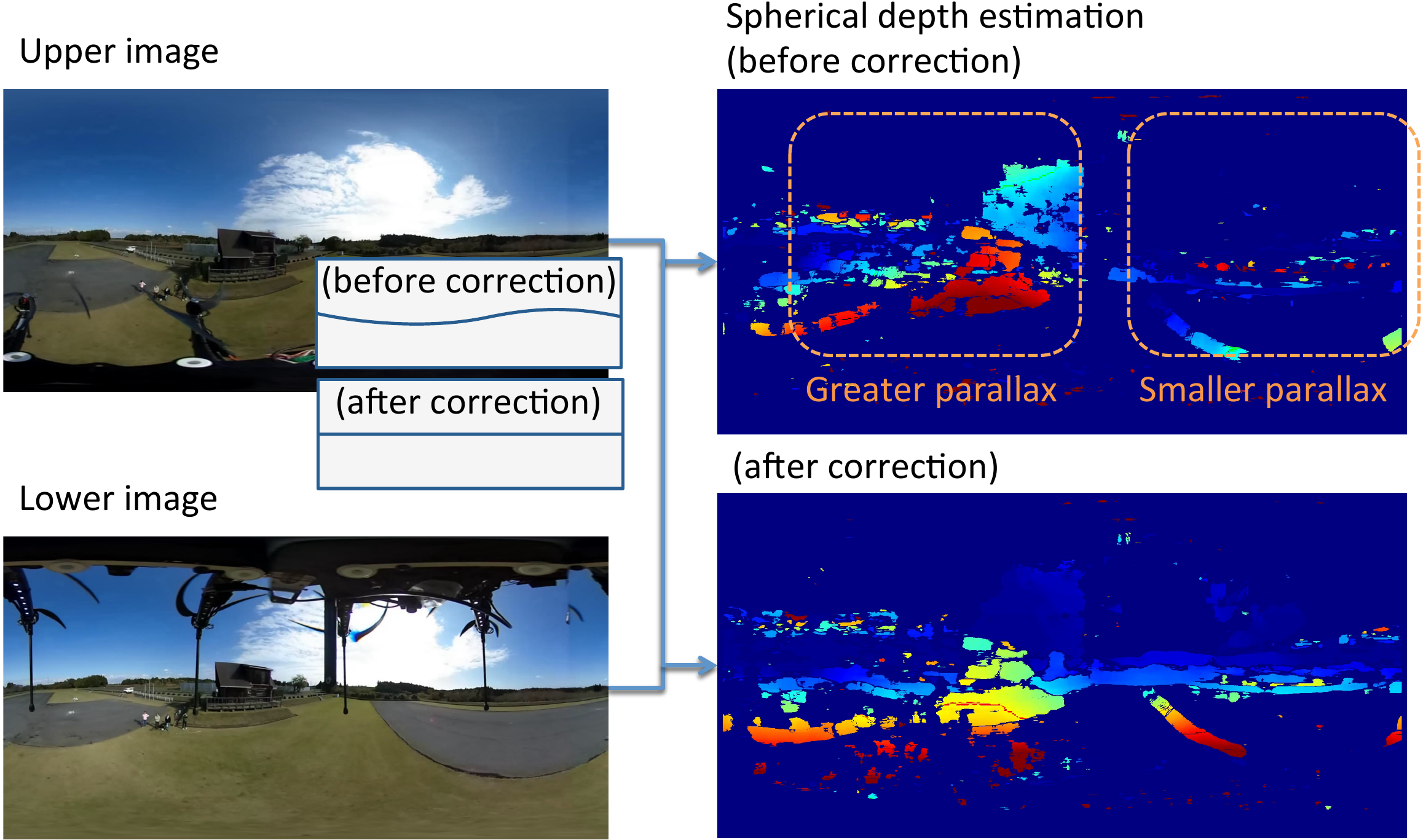}
  \caption{Depth estimation failure due to a camera misalignment.}
  \label{fig:binocular_drone_misalign}
\end{figure}

Figure \ref{fig:binocular_drone_misalign} shows an example of depth estimation failure due to an inaccurate correction of the top camera angles. In this case, because the rotation around the $x$ axis is not zero, the parallax for the left half of the equirectangular image becomes larger than the actual. This results in a smaller depth estimation in this area.
On the other hand, the parallax for the right half of the image becomes smaller than the actual (sometimes negative). As a result, the depth estimation in this area becomes greater than the actual.

\section{Monocular spherical stereo}

In this section, we extend the idea of binocular spherical stereo to retrieve spherical depth information from a monocular spherical video with motion parallax.
When we extract two adjacent frames from such video sequences, we can reconstruct spherical depth information for still or slow-moving objects.
Let $C_{t=t_0}$ and $C_{t=t_0 + \Delta t}$ the camera positions at $t=t_0$ and $t=t_0 + \Delta t$ respectively, then we can apply the binocular spherical depth estimation to this monocular spherical stereo problem.
We select the coordinate system for two viewpoints so that
\begin{enumerate}
\item $x$, $y$ and $z$ axes of the two viewpoints $C_{t=t_0}$ and $C_{t=t_0 + \Delta t}$ are in parallel.
\item Both $C_{t=t_0}$ and $C_{t=t_0 + \Delta t}$ share the $z$ axes\footnote{As the binocular depth estimation discussed in the previous section used the vertical parallax, we once converted the front view expression to a top view expression so that $z$ axis coincides the moving direction.}.
\end{enumerate}

In order for Condition 1 to be met, we have to cancel the roll, pitch and yaw rotation.
This problem is solved by n-point relative pose estimation\cite{hartley2003multiple}.
In order to stabilize the video, we followed the five-point relative pose estimation algorithm together with RANSAC proposed by Nist\'er\cite{nister2004efficient}. 
For Condition 2, we have to ensure the moving direction coincides the $z$ axis.
After the stabilization, we analyzed the optical flow on a spherical video for this purpose.

\subsection{Estimation of the moving direction}

\begin{figure}[ht]
  \centering
  \includegraphics[width=3.0in]{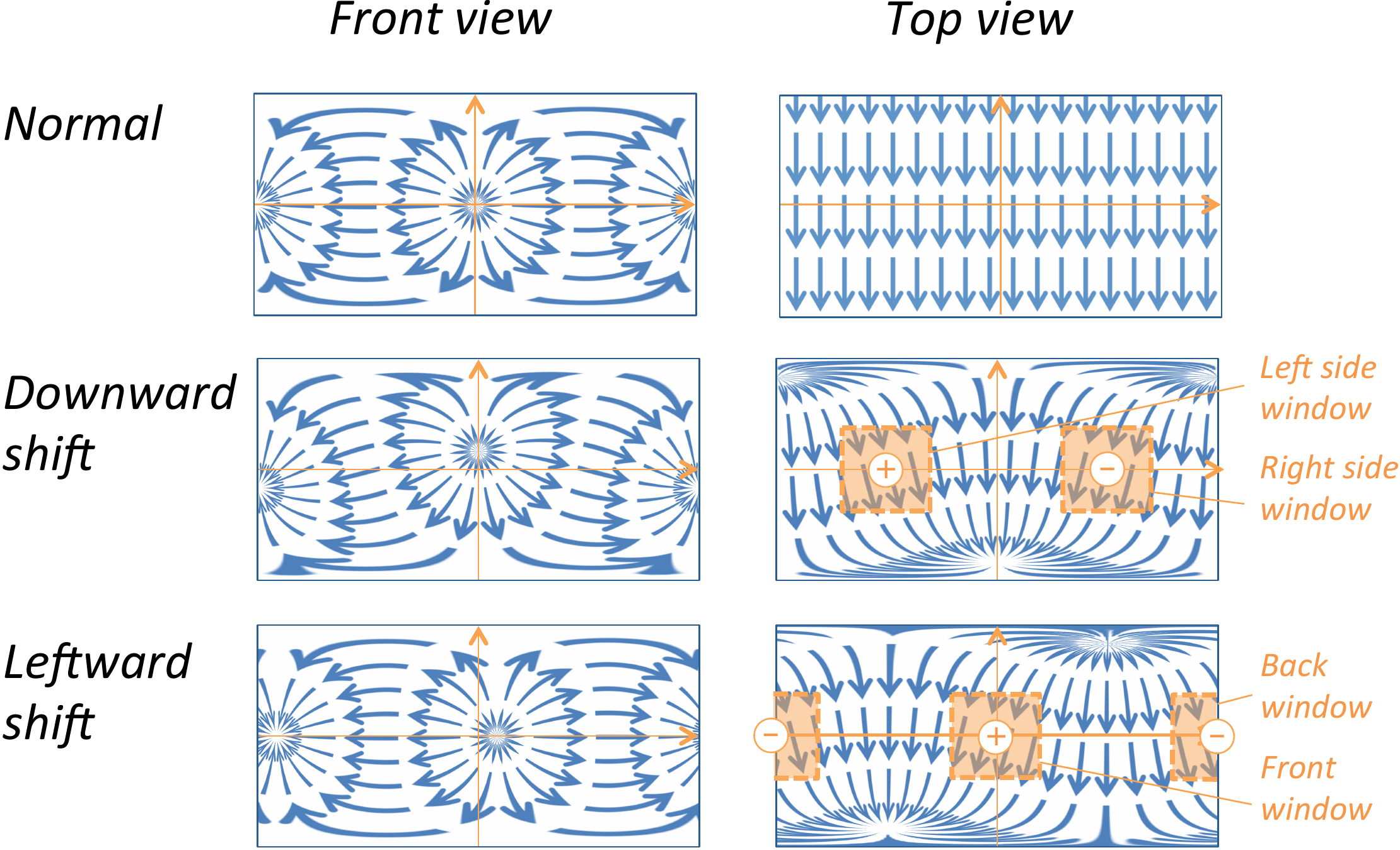}
  \caption{Optical flow on a spherical image caused by motion of viewpoint.
    When the vector source moves around the center, the optical flow in the top view is distorted.
  }
  \label{fig:md_estimation}
\end{figure}

In Figure \ref{fig:md_estimation}, the optical flow of the top view image is pointing straight downward at every location and the vectors are parallel when the vector source of the front view image is at the center.
When the position of the vector source moves upward
, $x$ components of the optical flow vectors in left side window becomes positive, while those in the right side window become negative.
Similarly, when the vector source moves to the right direction
, $x$ components in the front window becomes positive, while those in the back window become negative. Thus, we used the following calculation.
\begin{itemize}
\item Downward shift of the moving direction was estimated by calculating the ratio\\
    \small{\textbf{\texttt{median[union[$\mathbf{v}_x$ in left side window,\\ \hspace*{7em}$-\mathbf{v}_x$ in right side window]]}}}/
    \textbf{\texttt{median[union[$\mathbf{v}_y$ in left side window,\\ \hspace*{8em}$\mathbf{v}_y$ in right side window]]}}
\item Leftward shift of the moving direction was estimated by calculating the ratio\\
    \textbf{\texttt{median[union[$\mathbf{v}_x$ in front window,\\ \hspace*{7em}$-\mathbf{v}_x$ in back window]]}}/
    \textbf{\texttt{median[union[$\mathbf{v}_y$ in front window,\\ \hspace*{8em}$\mathbf{v}_y$ in back window]]}}
\end{itemize}

\begin{figure}[ht]
\centering
  \includegraphics[width=3.0in]{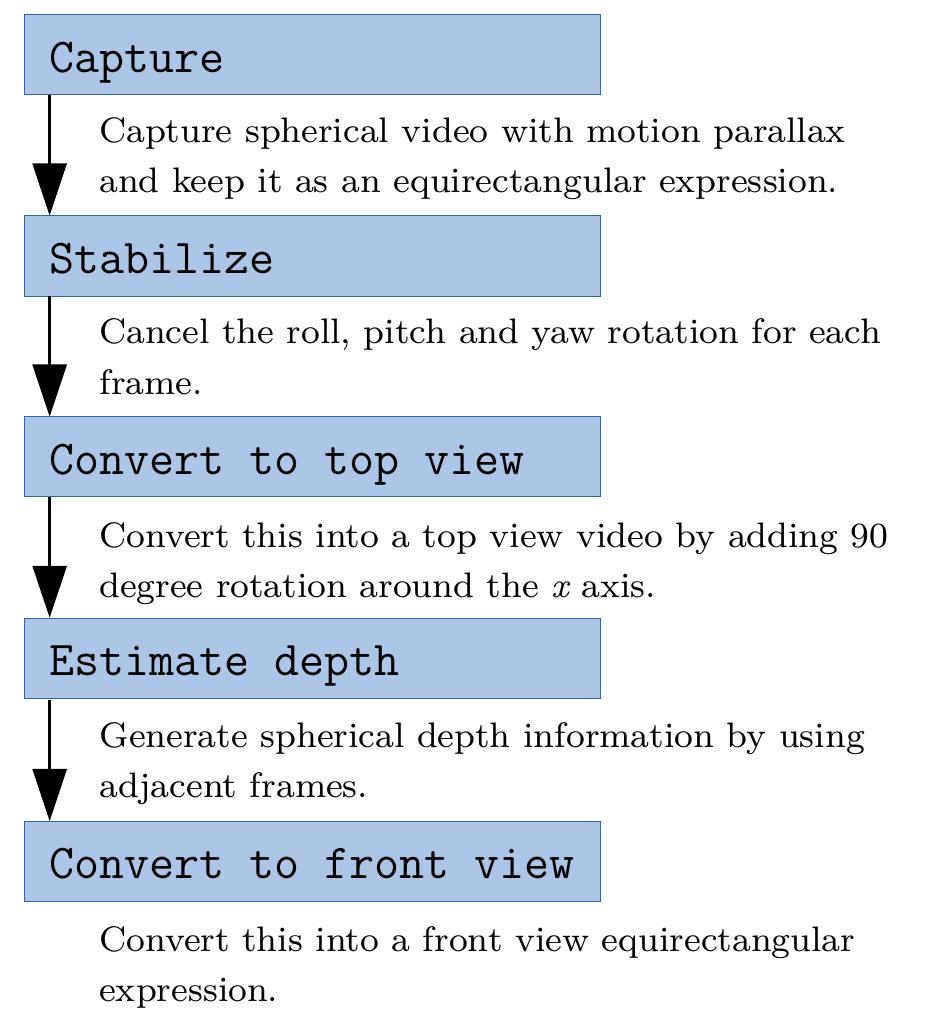}
  \caption{The processing pipeline of monocular spherical stereo.}
  \label{fig:monocular_spherical_stereo_pipeline}
\end{figure}

Figure \ref{fig:monocular_spherical_stereo_pipeline} shows the processing pipeline of monocular spherical stereo estimation.

\subsection{Experiments}
\label{subsec:mono_experiments}

\paragraph{Bike ride spherical video}

\begin{figure}[ht]
  \centering
  \includegraphics[width=1.2in]{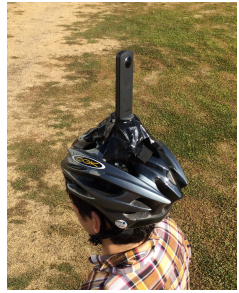}
  \caption{Ricoh THETA S attached on top of the bike helmet.}
  \label{fig:helmet_theta}
\end{figure}

We attached a Ricoh THETA S on top of the bike helmet and shot first-person spherical videos (Figure \ref{fig:helmet_theta}). The resolution of the equirectangular video was 1,920$\times$960. The video was recorded at 30FPS.
Since the typical speed of the bike was at 20 km/h, which corresponds to an inter-frame motion parallax of 20 cm, we used this length during the following calculation.

\begin{figure}[ht]
  \centering
  \includegraphics[width=3.0in]{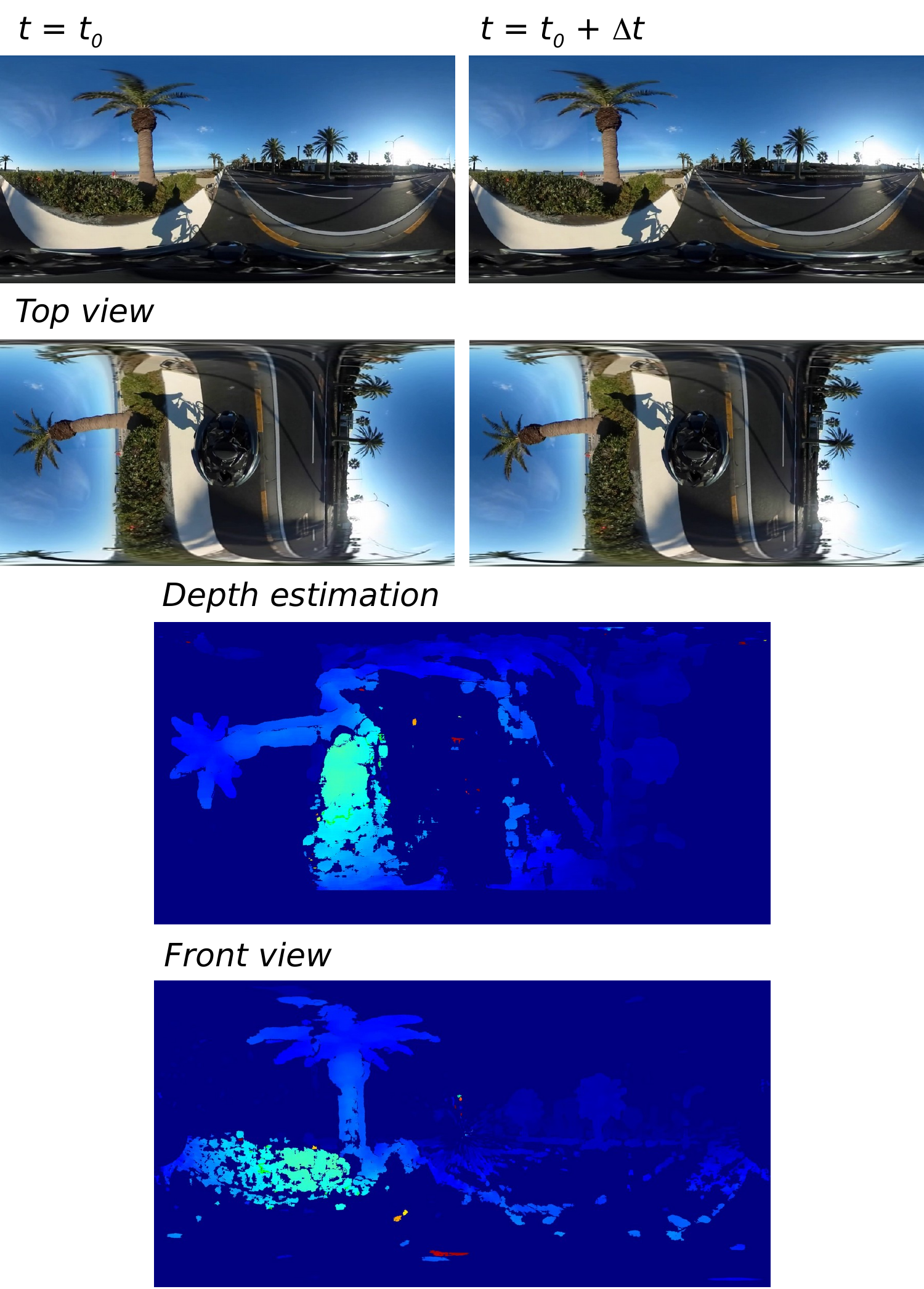}
  \caption{Depth estimation from a first-person spherical bike ride.
    We can observe the palm tree and the concrete wall. The farthest tree is located over 40 m from the viewpoint.
  }
  \label{fig:estdepth_result}
\end{figure}

Figure \ref{fig:estdepth_result} shows an example of the depth estimation from a first-person spherical video. We can observe the depth of buildings and palm trees. The farthest tree is located over 40 m from the viewpoint. We are not able to capture the depth of a moving car that is passing the bike from the stereo information.

\paragraph{Car driving spherical video}

\begin{figure}[ht]
  \centering
  \includegraphics[width=3.0in]{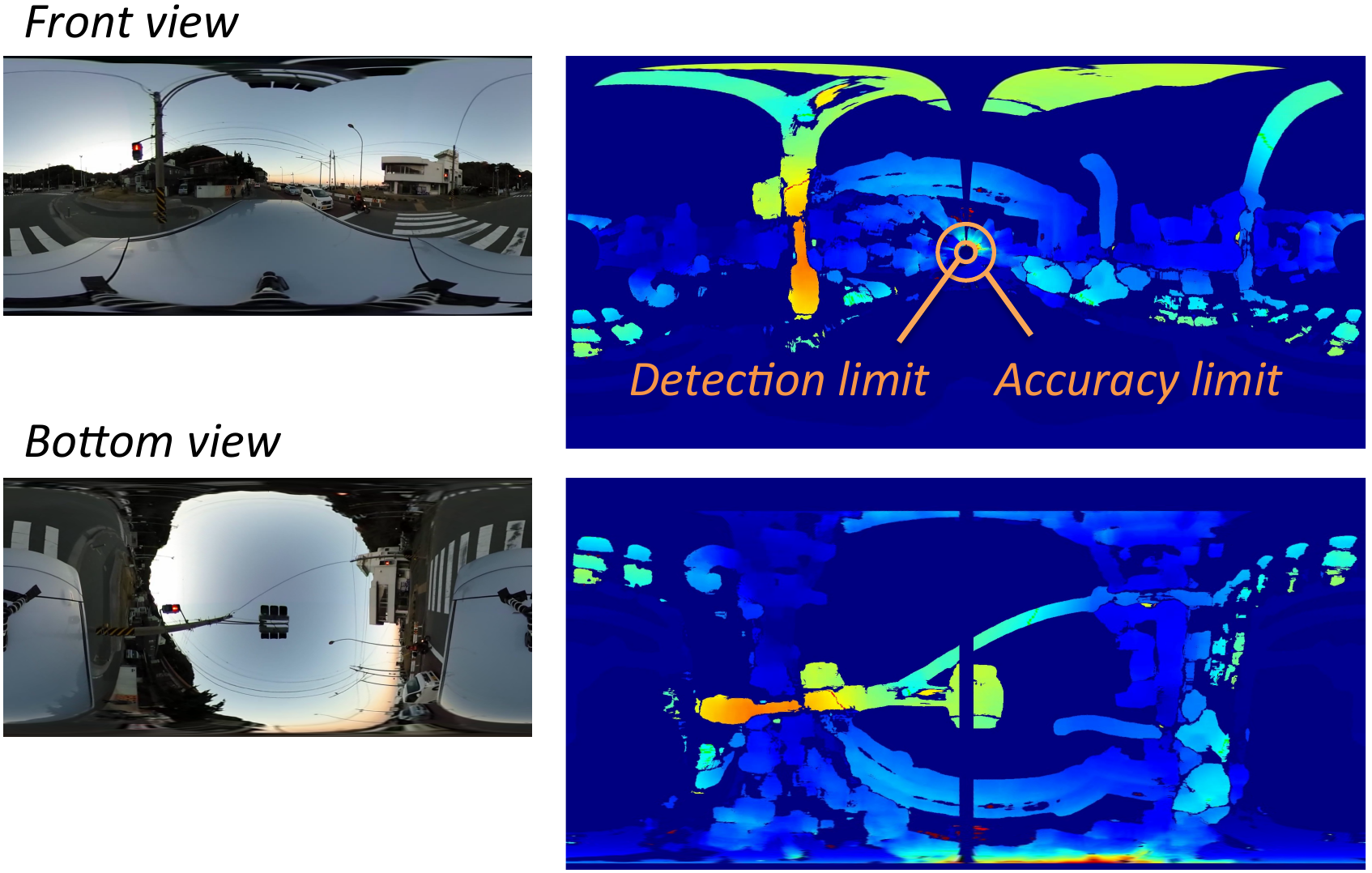}
  \caption{Angle coverage of a spherical depth estimation.
  }
  \label{fig:angle_coverage}
\end{figure}

We also recorded car driving videos.
This time, we used a Ricoh THETA V. The resolution of the equirectangular video was 3,840 x 1,920, and the video was recorded at 30FPS. The typical speed was at 30 km/h, which corresponds to an inter-frame motion parallax of 30 cm.
We evaluated the size of the blind spot in the front center (moving direction) and the back center (Figure \ref{fig:angle_coverage} front view).
The front and back moving directions correspond to the north and south high latitude area in the binocular spherical stereo described in Section \ref{sec:binocular_spherical_stereo}.

Let $\psi$ the half of the visual angle of the blind spot, the ratio of the blind spot surface area is formulated as follows:
\begin{equation}
  A_{blind}(\psi) = \left( 2 \pi \int_0^{\psi} \! \sin\psi \,  \mathrm{d}\psi \right) \Big/ 2 \pi = 1 - \cos \psi
\end{equation}
Detection limit is defined by the stereo block matching margin. Using the coordinate system for the binocular spherical stereo, we evaluated this limit at around north latitude $85^{\circ}$. This results in the sphere coverage ratio of $1 - A_{blind} = 99.56\%$.
On the other hand, regarding the accuracy limit, below which we could obtain a stable depth information, we evaluated this limit at around north latitude $77^{\circ}$. This results in the sphere coverage ratio of $1 - A_{blind} = 97.6\%$.
In the bottom view, We also found that we can capture the depth of the traffic signal above the viewpoint.

\paragraph{Environment 3D reconstruction}

We further converted the spherical depth data into point cloud data using the method mentioned in Section \ref{subsec:binocular_results}. Again, the number of disparities in stereo matching was 96.
We calculated and accumulated the point cloud for every 30 frames and merged the data into one 3D reconstruction. In order to merge the data with appropriate offset, we estimated the speed of the vehicle and assumed it was constant as well as the moving direction. If the speed is varying, we will obtain the 3D reconstruction result with varying scale. Also if the moving direction is varying, we will obtain the result with varying directions. In both cases we can rotate and translate the 3D data to match the result from the previous frame(s) by utilizing accurate registration algorithms such as iterative closest point (ICP) registration\cite{besl1992method}.

\begin{figure}[ht]
  \centering
  \includegraphics[width=2.5in]{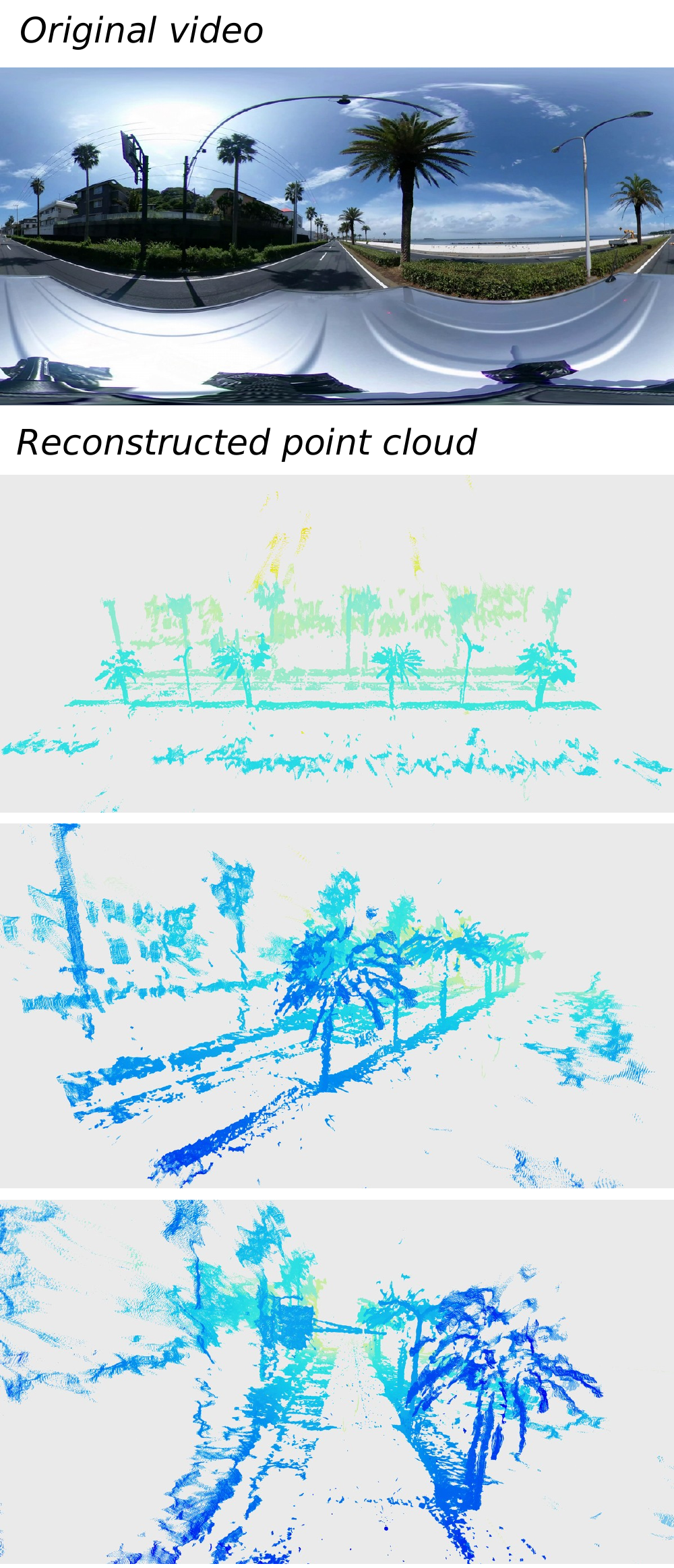}
  \caption{Example of reconstructed point cloud. Accumulation of eight captures in eight seconds. We can clearly see the row of palm trees.}
  \label{fig:pointcloud_ex_1}
\end{figure}

\begin{figure}[ht]
  \centering
  \includegraphics[width=2.5in]{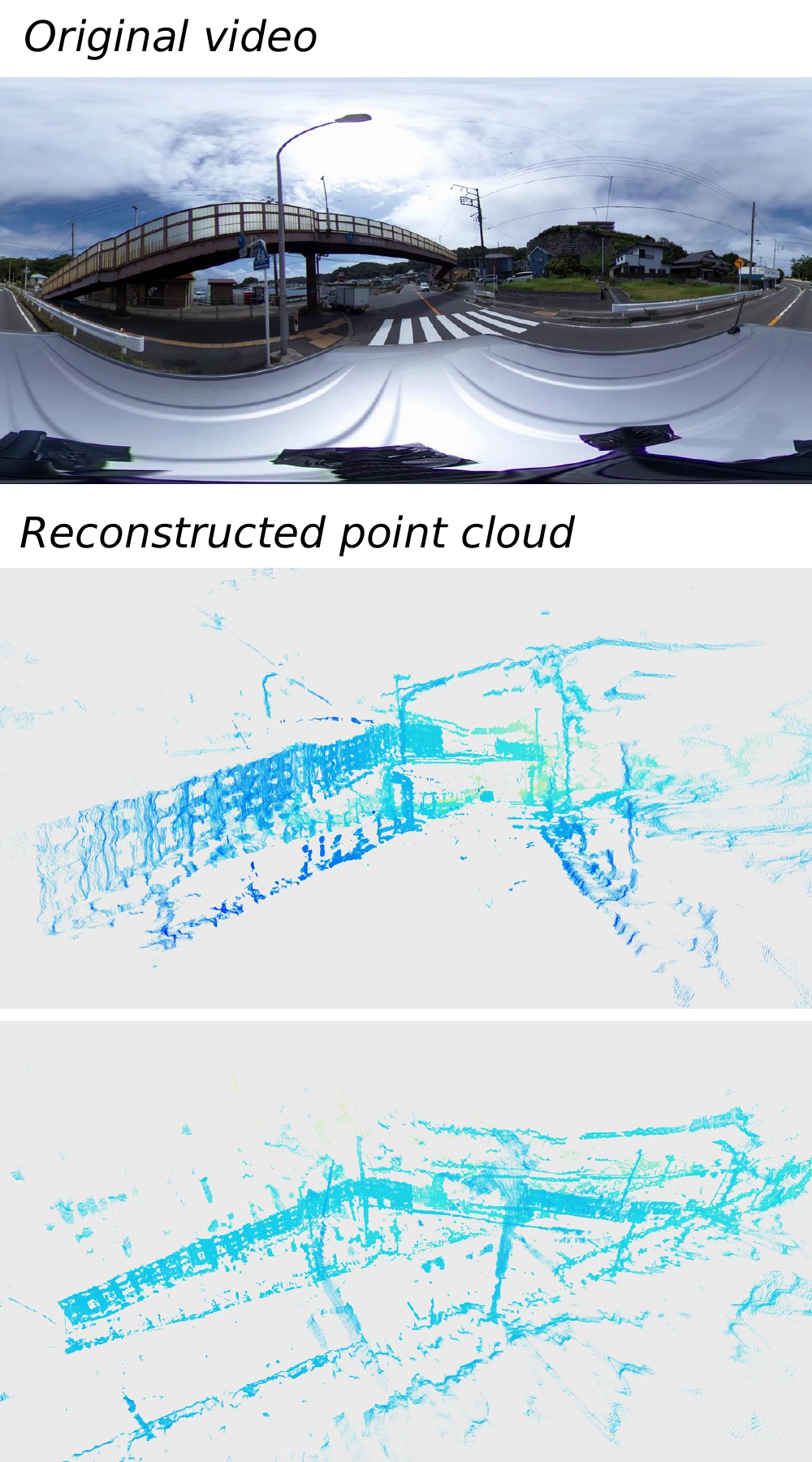}
  \caption{Example of reconstructed point cloud. Accumulation of eight captures in eight seconds. We can clearly see the structure of the pedestrian slope.}
  \label{fig:pointcloud_ex_2}
\end{figure}

Figure \ref{fig:pointcloud_ex_1} and \ref{fig:pointcloud_ex_2} show the examples of the 3D reconstruction of the environments.
Using Open3D point cloud visualization, we can clearly observe the 3D structure of a row of palm trees, traffic signs and a pedestrian slope.
Since we have 3D environmental data, we can place virtual cameras at arbitrary positions.
Creating such bird-view graphics is useful for many applications including 1) camera work planning and previsualization for filmmakers and sports producers, 2) automatic robot or vehicle navigation and 3) geographical information services.

\paragraph{Application to 360 videos on the Internet}

\begin{figure}[ht]
  \centering
  \includegraphics[width=3.0in]{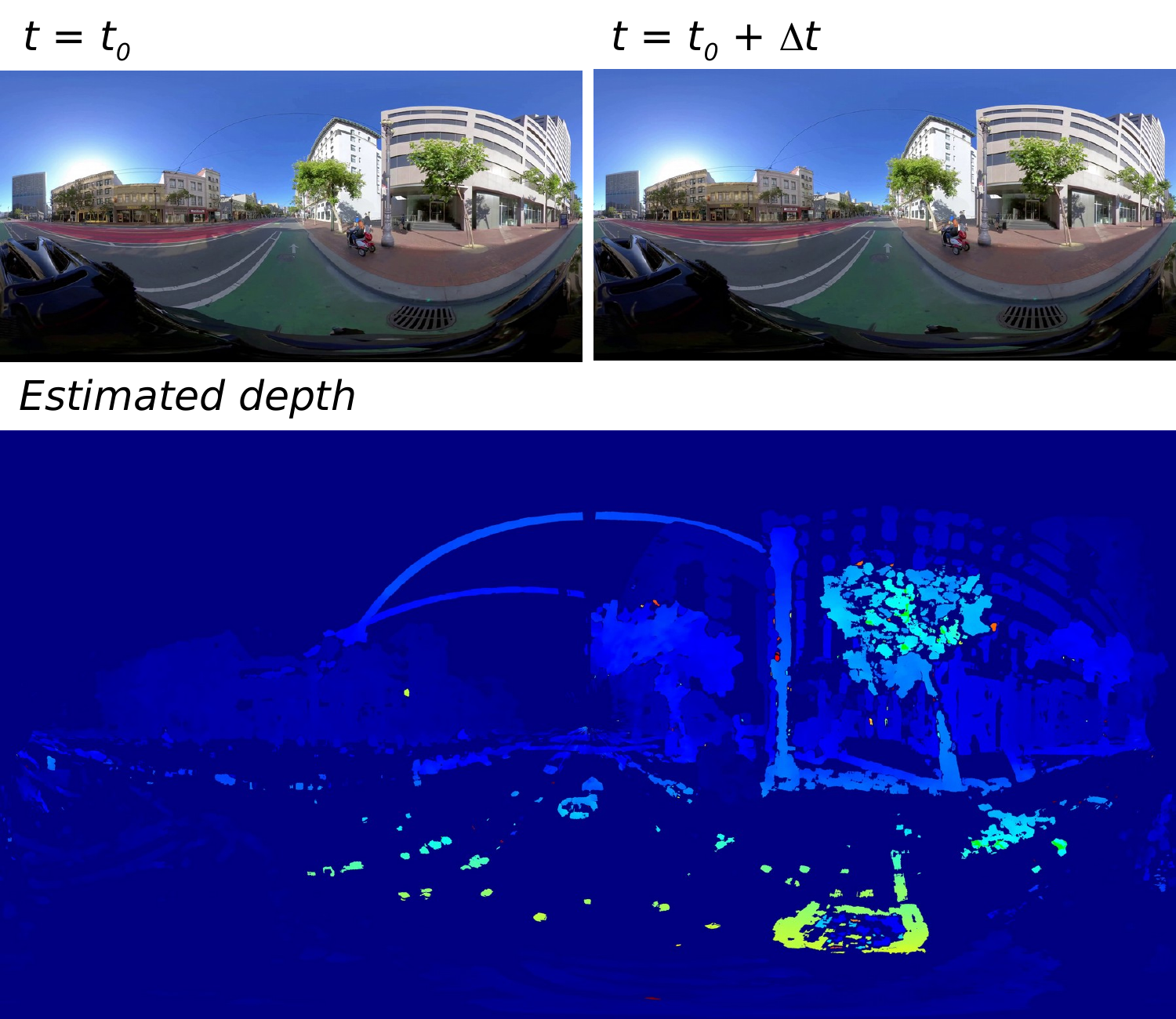}
  \caption{Depth estimation for a first-person bike ride video on YouTube. The resolution was 3,840$\times$1,920.}
  \label{fig:estdepth_result3}
\end{figure}

We applied this method to a first-person spherical video on the Internet.
With equirectangular videos, we don't need to estimate the field of view (FOV) of a footage. Often, we can easily determine the horizontal line and the zenith position.
Figure \ref{fig:estdepth_result3} shows an example of depth estimation from a first-person bike ride footage on YouTube. We can clearly see the structure of the trees and the buildings.

\paragraph{Application to aerial videos}

\begin{figure}[ht]
  \centering
  \includegraphics[width=3.0in]{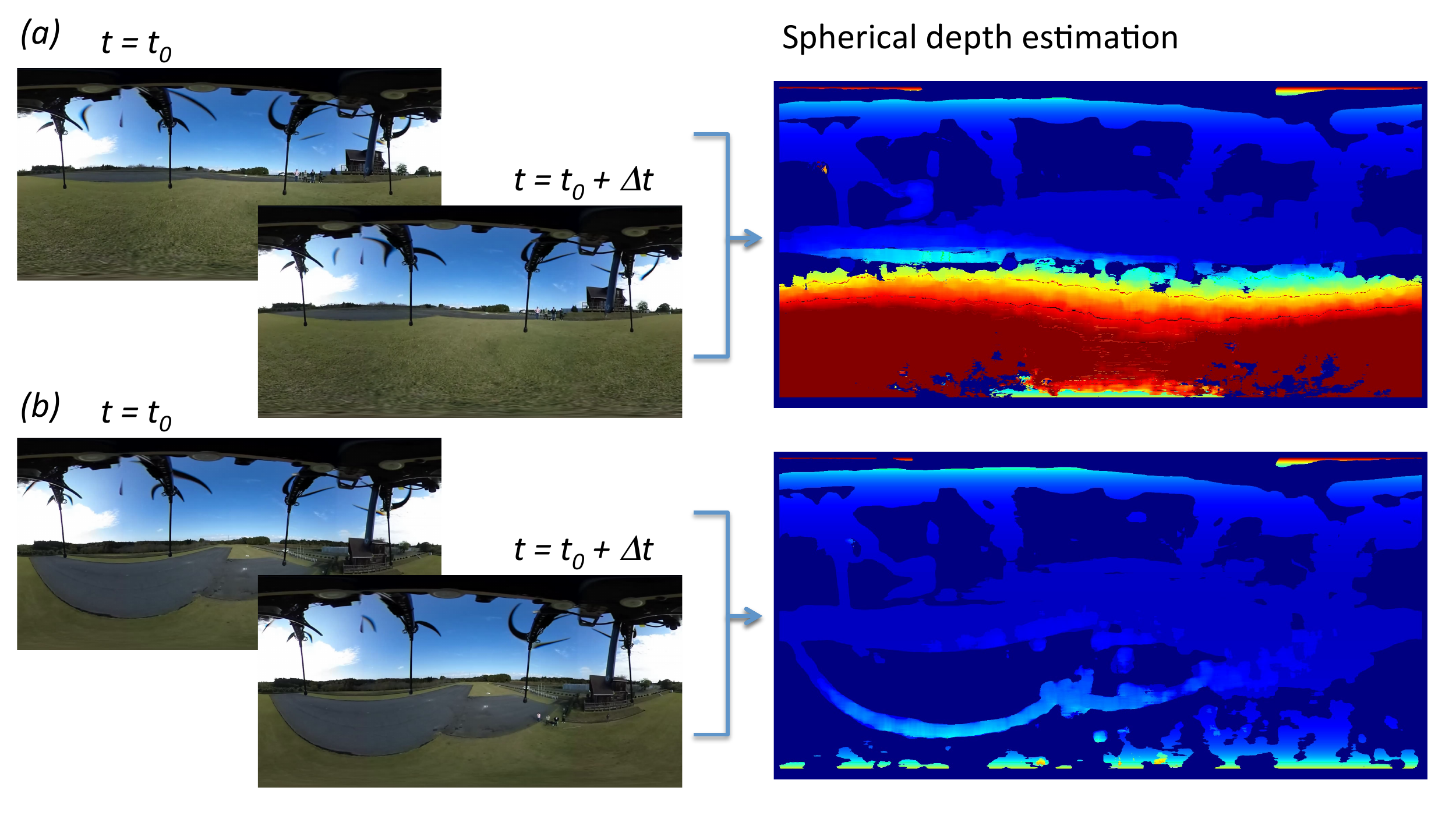}
  \caption{Spherical depth estimation from a monocular spherical camera attached to a drone at different heights.
    In this sequence of take-off, the motion parallax was vertical.}
  \label{fig:monocular_drone}
\end{figure}

It is essential to sense the entirely spherical environment in an automatic navigation of drones.
While we already presented the application of binocular spherical stereo to aerial videos in Section \ref{sec:binocular_spherical_stereo},
we also tried the monocular method (Figure \ref{fig:monocular_drone}).
Spherical depth information was calculated from a take-off scene shot by a Ricoh THETA S attached below a drone. This time, the motion parallax was vertical.
At lower positions we can observe the ground clearly (a), while at higher positions we can barely observe the edge of the runway (b).
We have implemented a prototype of an obstacle detection system by utilizing this algorithm running on an NVIDIA Jetson.

\section{Discussions}
In this section, we discuss the advantages and the challenges of monocular spherical stereo in comparison to binocular method.

\subsection{Monocular method advantages}

\paragraph{Consistent camera characteristics}
Because the images are captured with a single camera, we do not need to consider the difference in the characteristics of the sensors. For example, we set the camera capture mode to automatic during the outdoor shooting so that it accommodates dynamic range of brightness. When we use binocular spherical stereo for capturing, we need to watch if the characteristics of the two cameras are exactly the same. 
\paragraph{Less visual obstacles}
To use binocular spherical stereo for capturing, we need to fix two cameras, creating at least two blind spots and 
the occlusion caused by the adjacent fisheye lens narrows the practical angles for a stereo observation.
However, when we use the monocular method, we have only one blind spot. 
This is clear when we compare the monocular capture with binocular spherical stereo. 
\paragraph{No need for synchronization}
When we use a monocular method, we do not need to synchronize two cameras by definition. Instead, a constant 1/30 second gap (when the frame rate is 30FPS) is guaranteed between two images.

\subsection{Monocular method challenges}

\paragraph{Less accuracy for moving objects}
Clearly, the ability to detect a moving object is limited.
In the bike ride video, we can observe that we cannot capture the depth of a car that is passing the bike, while we can detect an oncoming car.
This is because the stereo problem for a passing car contains ill-formed, divergence condition,
while the parallax always increases for an oncoming car.


\section{Conclusion and future work}

We described and demonstrated a method to capture entirely spherical depth information and reconstruct the 3D structures (point cloud) using two frames from a single spherical video.
After illustrating a spherical depth information retrieval using two spherical cameras, we demonstrated monocular spherical stereo by using stabilized video footage with motion parallax.
Experiments demonstrated that the depth information covered up to 97\% (nearly entirely spherical) of the sphere in solid angle. By comparing the 3D reconstruction with the actual floor plan, we found that the maximum error in the dimensions obtained from the point cloud is less than 5\% for a certain range.
With two cameras separated by 35 cm vertically, we were able to detect an object located over 100 m from the viewpoint.

Furthermore, we reconstructed the 3D point cloud data using the spherical depth data from monocular spherical video with motion parallax.
By accumulating 3D data from several captures, we confirmed the structures such as roads, buildings, pedestrian slope and palm trees can be clearly observed. 
At a speed of 30 km/h, we were able to detect an object located over 30 m from the moving camera. We applied the method to bike ride, car driving and aerial spherical videos as well as 360 videos on the Internet.

Future works include automated accumulation of 3D data.
As discussed in Section \ref{subsec:mono_experiments}, we are able to reconstruct a wider area by accumulating 3D data without assuming the speed and the moving direction of the viewpoint are constant. We can rotate and translate the most current 3D data to match the accumulated 3D data by using accurate registration algorithms such as ICP. 
Developing an integrated tool that simply converts the first-person spherical video into wide-range 3D structure data is also included in the future works.
This can be further integrated into previsualization and location hunting tools for creators.


\appendix

\section{Distance rectification for high latitude areas}
\label{sec:apparent_reduction}
\begin{figure}[ht]
  \centering
  \includegraphics[width=2.0in]{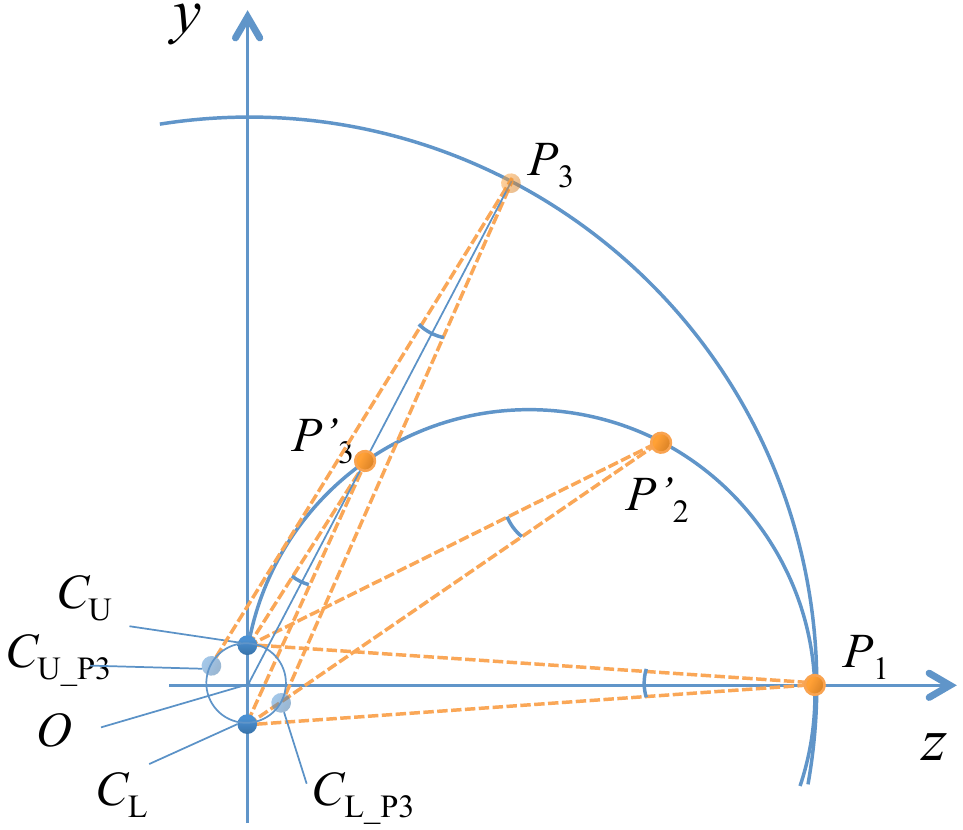}
  \caption{Latitude based rectification of the distance.}
  \label{fig:high_latitude_rectify}
\end{figure}

In Figure \ref{fig:high_latitude_rectify}, two cameras are plotted as $C_U$ and $C_L$. The distance $\|C_U C_L\|$ is the interpupilar distance for an object that is on the $y = 0$ plane.
However, the distances of the points in the high latitude area are underestimated because of the apparent reduction of the interpupilar distance.
For example, $\|O P_3\|$ equals to $\|O P_1\|$ if the interpupilar distance doesn't vary as the latitude changes ($\|C_{U\textunderscore P3} C^\prime_{L\textunderscore P3}\| = \|C_U C_L\|$). In reality, the apparent distance of $P_3$ is reduced:

\begin{equation}
 \|O P^\prime_3\| = \cos \angle P_3 O P_1 \|O P_1\|
\end{equation}

By the inscribed angle theorem, regarding the vertical visual angle of Point $P_1$, $P_2$ and $P_3$, the formula
\begin{equation}
  \angle C_U P_1 C_L = \angle C_U P_2 C_L = \angle C_U P_3 C_L
\end{equation}  
is established when $P_1$, $P_2$, $P_3$, $C_U$ and $C_L$ are on the same circle.

As a result, we have to rectify the distance of an object based on the latitude $\phi$ of the incoming light of the object.
\begin{equation}
  D_{REAL} \simeq \frac{D_{APPARENT}}{\cos \phi}
\end{equation}

\section{Roll, pitch and yaw correction of an equirectangular image}
\label{sec:rpy_correct}

In Figure \ref{fig:binocular_cameras}, the position of $P$ is described as:
\begin{equation}
  \mathbf{v_P} = \left(
  \begin{matrix}
    \cos \phi \sin \theta \\
    \sin \phi \\
    \cos \phi \cos \theta
  \end{matrix}
  \right)
\end{equation}

Then, the rotations around $x$, $y$ and $z$ axes are represented as the followings respectively:
\begin{eqnarray}
  \mathbf{R_x(\alpha)} & = &
  \left(
  \begin{array}{ccc}
    1 & 0 & 0 \\
    0 & \cos \alpha & -\sin \alpha \\
    0 & \sin \alpha & \cos \alpha \\
  \end{array}
  \right)\\
  \mathbf{R_y(\beta)} & = &
  \left(
  \begin{array}{ccc}
    \cos \beta & 0 & \sin \beta \\
    0 & 1 & 0 \\
    -\sin \beta & 0 & \cos \beta \\
  \end{array}
  \right)\\
  \mathbf{R_z(\gamma)} & = &
  \left(
  \begin{array}{ccc}
    \cos \gamma & -\sin \gamma & 0 \\
    \sin \gamma & \cos \gamma & 0 \\
    0 & 0 & 1 \\
  \end{array}
  \right)
\end{eqnarray}

\paragraph{General roll, pitch and yaw correction}
We can calculate the matrix that represents the sequential rotation around the $x$, $y$ and $z$. When $\alpha$, $\beta$ and $\gamma$ are small enough, we can use approximations again to obtain the following:
\begin{equation}
  \mathbf{R_{xyz}(\alpha, \beta, \gamma)} =
  \left(
  \begin{array}{ccc}
    1 & -\gamma & \beta \\
    \gamma & 1 & -\alpha \\
    -\beta & \alpha & 1 \\
  \end{array}
  \right)
\end{equation}
Although this is an approximation applicable only if the angles are small enough, the result doesn't depend on the order of the rotations.

\bibliographystyle{ieee_fullname}
\bibliography{sphericalstereo}

\begin{thebibliography}{10}\itemsep=-1pt

\bibitem{anguelov2010google}
Dragomir Anguelov, Carole Dulong, Daniel Filip, Christian Frueh, St{\'e}phane
  Lafon, Richard Lyon, Abhijit Ogale, Luc Vincent, and Josh Weaver.
\newblock Google street view: Capturing the world at street level.
\newblock {\em Computer}, (6):32--38, 2010.

\bibitem{bartczak2007extraction}
Bogumil Bartczak, Kevin Koeser, Felix Woelk, and Reinhard Koch.
\newblock Extraction of 3d freeform surfaces as visual landmarks for real-time
  tracking.
\newblock {\em Journal of Real-Time Image Processing}, 2(2-3):81--101, 2007.

\bibitem{besl1992method}
Paul~J Besl and Neil~D McKay.
\newblock Method for registration of 3-d shapes.
\newblock In {\em Sensor fusion IV: control paradigms and data structures},
  volume 1611, pages 586--606. International Society for Optics and Photonics,
  1992.

\bibitem{bodington2015rendering}
Dash Bodington, Jayant Thatte, and Matthew Hu.
\newblock Rendering of stereoscopic 360 views from spherical image pairs.
\newblock 2015.

\bibitem{broxton2020immersive}
Michael Broxton, John Flynn, Ryan Overbeck, Daniel Erickson, Peter Hedman,
  Matthew Duvall, Jason Dourgarian, Jay Busch, Matt Whalen, and Paul Debevec.
\newblock Immersive light field video with a layered mesh representation.
\newblock {\em ACM Transactions on Graphics (TOG)}, 39(4):86--1, 2020.

\bibitem{hartley2003multiple}
Richard Hartley and Andrew Zisserman.
\newblock {\em Multiple view geometry in computer vision}.
\newblock Cambridge university press, 2003.

\bibitem{huang20176}
Jingwei Huang, Zhili Chen, Duygu Ceylan, and Hailin Jin.
\newblock 6-dof vr videos with a single 360-camera.
\newblock In {\em 2017 IEEE Virtual Reality (VR)}, pages 37--44. IEEE, 2017.

\bibitem{kim20133d}
Hansung Kim and Adrian Hilton.
\newblock 3d scene reconstruction from multiple spherical stereo pairs.
\newblock {\em International journal of computer vision}, 104(1):94--116, 2013.

\bibitem{li2006real}
Shigang Li.
\newblock Real-time spherical stereo.
\newblock In {\em Pattern Recognition, 2006. ICPR 2006. 18th International
  Conference on}, volume~3, pages 1046--1049. IEEE, 2006.

\bibitem{li2008binocular}
Shigang Li.
\newblock Binocular spherical stereo.
\newblock {\em Intelligent Transportation Systems, IEEE Transactions on},
  9(4):589--600, 2008.

\bibitem{lin2003high}
ShihSchon Lin and Ruzena Bajcsy.
\newblock High resolution catadioptric omni-directional stereo sensor for robot
  vision.
\newblock In {\em Robotics and Automation, 2003. Proceedings. ICRA'03. IEEE
  International Conference on}, volume~2, pages 1694--1699. IEEE, 2003.

\bibitem{ma20153d}
Chuiwen Ma, Liang Shi, Hanlu Huang, and Mengyuan Yan.
\newblock 3d reconstruction from full-view fisheye camera.
\newblock {\em arXiv preprint arXiv:1506.06273}, 2015.

\bibitem{munteanu2014visual}
O Munteanu, R Pronk, and Arnoud Visser.
\newblock Visual odometry with the ricoh theta.
\newblock {\em Project Report, Universiteit van Amsterdam, February}, 2014.

\bibitem{nister2004efficient}
David Nist{\'e}r.
\newblock An efficient solution to the five-point relative pose problem.
\newblock {\em Pattern Analysis and Machine Intelligence, IEEE Transactions
  on}, 26(6):756--770, 2004.

\bibitem{scaramuzza2008appearance}
Davide Scaramuzza and Roland Siegwart.
\newblock Appearance-guided monocular omnidirectional visual odometry for
  outdoor ground vehicles.
\newblock {\em Robotics, IEEE Transactions on}, 24(5):1015--1026, 2008.

\bibitem{shimamura2000construction}
Jun Shimamura, Naokazu Yokoya, Haruo Takemura, and Kazumasa Yamazawa.
\newblock Construction of an immersive mixed environment using an
  omnidirectional stereo image sensor.
\newblock In {\em Omnidirectional Vision, 2000. Proceedings. IEEE Workshop on},
  pages 62--69. IEEE, 2000.

\bibitem{tanaka2005tornado}
Kenji Tanaka and Susumu Tachi.
\newblock Tornado: omnistereo video imaging with rotating optics.
\newblock {\em Visualization and Computer Graphics, IEEE Transactions on},
  11(6):614--625, 2005.

\bibitem{tanhashi2001acquisition}
H Tanhashi, Daisuke Shimada, Kazuhiko Yamamoto, and Yoshinori Niwa.
\newblock Acquisition of three-dimensional information in a real environment by
  using the stereo omni-directional system (sos).
\newblock In {\em 3-D Digital Imaging and Modeling, 2001. Proceedings. Third
  International Conference on}, pages 365--371. IEEE, 2001.

\bibitem{zhou2018open3d}
Qian-Yi Zhou, Jaesik Park, and Vladlen Koltun.
\newblock Open3d: A modern library for 3d data processing.
\newblock {\em arXiv preprint arXiv:1801.09847}, 2018.

\bibitem{zioulis2018omnidepth}
Nikolaos Zioulis, Antonis Karakottas, Dimitrios Zarpalas, and Petros Daras.
\newblock Omnidepth: Dense depth estimation for indoors spherical panoramas.
\newblock In {\em Proceedings of the European Conference on Computer Vision
  (ECCV)}, pages 448--465, 2018.

\end{thebibliography}

\end{document}